\documentclass[conference]{IEEEtran}
\IEEEoverridecommandlockouts
\usepackage{cite}
\usepackage{amsmath,amssymb,amsfonts}
\usepackage{algorithmic}
\usepackage{graphicx}
\usepackage{textcomp}
\usepackage{xcolor}
\usepackage{hyperref}
\usepackage{multirow}
\usepackage{subfig}

\DeclareMathOperator*{\argmin}{\arg\!\min}
\DeclareMathOperator*{\argmax}{\arg\!\max}

\def\BibTeX{{\rm B\kern-.05em{\sc i\kern-.025em b}\kern-.08em
    T\kern-.1667em\lower.7ex\hbox{E}\kern-.125emX}}
\begin{document}

\title{Decentralized Motion Planning for Multi-Robot Navigation using Deep Reinforcement Learning\\
}

\author{\IEEEauthorblockN{K. Sivanathan, Tanmay Samak, Chinmay Samak}
\IEEEauthorblockA{\textit{Autonomous Systems Lab, Mechatronics Engineering Dept.} \\
\textit{SRM Institute of Science and Technology}\\
Tamil Nadu, India \\
\tt{\{\href{mailto:sivanatk@srmist.edu.in}{sivanatk}, \href{mailto:tv4813@srmist.edu.in}{tv4813}, \href{mailto:cv4703@srmist.edu.in}{cv4703}\}@srmist.edu.in}}
\and
\IEEEauthorblockN{B. K. Vinayagam}
\IEEEauthorblockA{\textit{Mechatronics Engineering Dept. (Former Professor)} \\
\textit{SRM Institute of Science and Technology}\\
Tamil Nadu, India \\
\tt{\href{mailto:bkvei23@gmail.com}{bkvei23@gmail.com}}}}

\maketitle

\begin{abstract}
This work presents a decentralized motion planning framework for addressing the task of multi-robot navigation using deep reinforcement learning. A custom simulator was developed in order to experimentally investigate the navigation problem of 4 cooperative non-holonomic robots sharing limited state information with each other in 3 different settings. The notion of decentralized motion planning with common and shared policy learning was adopted, which allowed robust training and testing of this approach in a stochastic environment since the agents were mutually independent and exhibited asynchronous motion behavior. The task was further aggravated by providing the agents with a sparse observation space and requiring them to generate continuous action commands so as to efficiently, yet safely navigate to their respective goal locations, while avoiding collisions with other dynamic peers and static obstacles at all times. The experimental results are reported in terms of quantitative measures and qualitative remarks for both training and deployment phases.
\end{abstract}


\section{Introduction}\label{Introduction}
The problem of online motion planning for a fleet of autonomous mobile robots has been explored over decades employing a number of approaches for applications including search and rescue missions\cite{queralta2020collaborative}, collaborative material transfer and construction\cite{augugliaro2014flight}, warehouse management\cite{ocado}, connected autonomous vehicles\cite{loke2019cooperative} and even aesthetic performances\cite{augugliaro2013dance}.

In the context of multi-robot motion planning, the algorithms can be essentially implemented in a centralized\cite{luna2011efficient} or decentralized\cite{ayanian2012decentralized} manner. While the prior approach employs a central server for state estimation, prediction and motion planning of all the robots, the later one employs independent on-board resources such that each robot plans its own motion considering the states of others (either through exteroceptive sensing for non-cooperative agents or inter-robot communication for cooperative ones). Although centralized implementations are optimal and safe, their performance heavily degrades when scalability and robustness are desired, especially in highly dynamic environments. This work therefore focuses on decentralized implementations aimed at optimizing the local trajectories temporally and spatially.

\begin{figure}[htbp]
	\centerline{\includegraphics[width=0.5\textwidth]{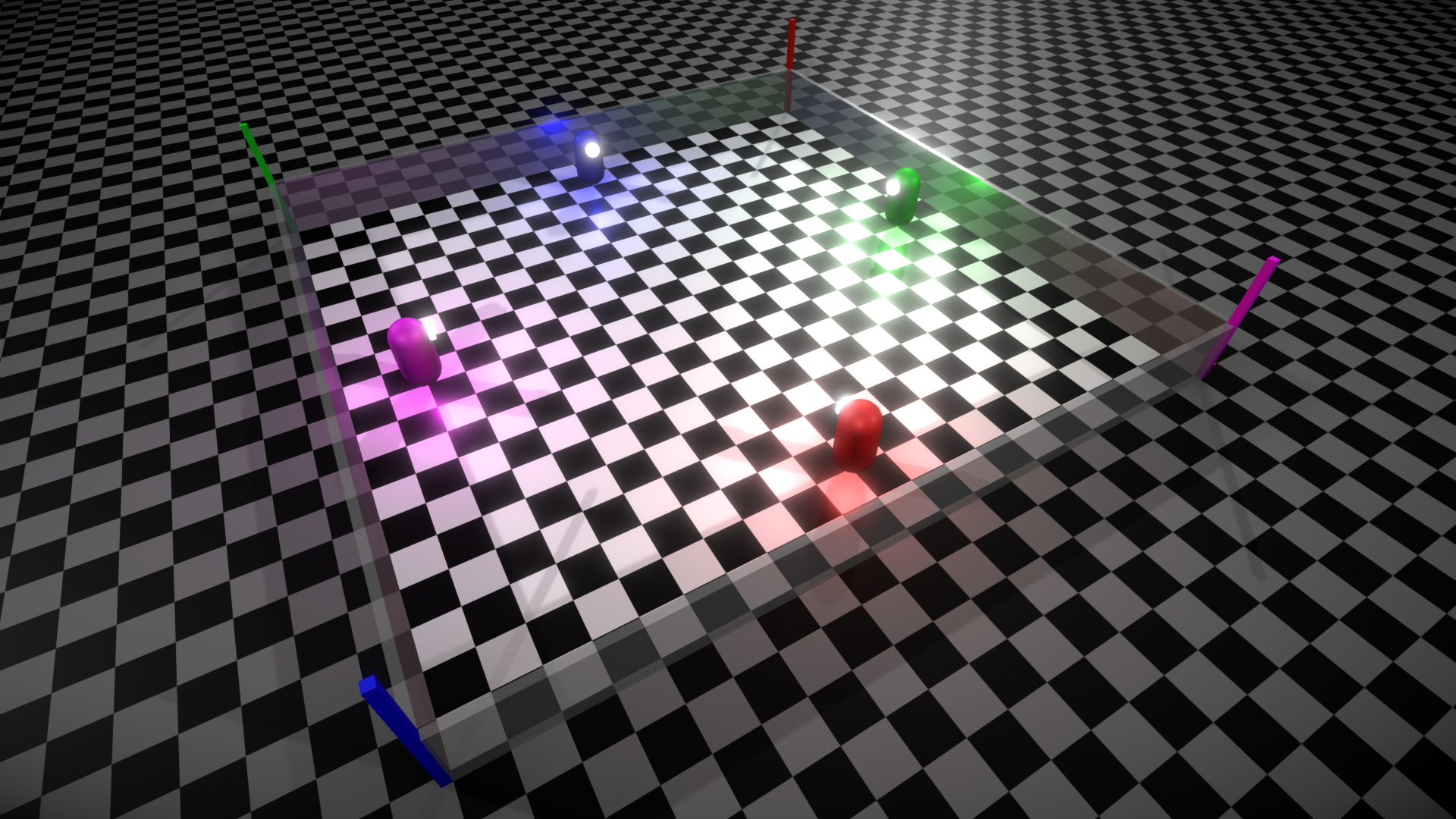}}
	\caption{MARL Simulator running a sample multi-robot navigation scenario of 4 agents within a square arena.}
	\label{fig: MARL Simulator}
\end{figure}

Principally, a decentralized motion planning framework can be either implemented using Classical Optimization (CO) based methods or Artificial Intelligence (AI) based methods.

The CO based methods\cite{berg2008reciprocal, berg2011reciprocal, daniel2012multi-robot} may be further categorized as constrained or unconstrained. Theoretically, the constrained optimization methods provide safety guarantee, but take longer travel time and become intractable in higher dimensions. The unconstrained optimization methods, on the other hand, employ either no constraints or a limited number of soft constraints, and tend to be faster in execution and travel time but less safe \cite{kandhasamy2020scalable}. Nonetheless, CO based methods do not provide safety guarantee in practice owing to the fact that multi-robot motion planning is a rather complex problem demanding highly nonlinear models to be solved in real-time with persistent model and measurement uncertainties.

The AI based methods, more particularly the learning based methods, address this issue by making use of artificial neural networks as non-linear function approximators, which inherently learn to cope with the uncertainties and enable real-time execution in the inference stage. Deep Imitation Learning (DIL) methods are well suited for single-agent systems such as autonomous vehicles \cite{samak2020robust} but are seldom applied directly to multi-agent systems owing to labeled-data requirement of the permutative and complex interactions. Deep Reinforcement Learning (DRL) methods \cite{chen2016decentralized, chen2018socially, everett2018motion, semnani2020multiagent, long2018optimally, aradi2020survey, wang2020mrcdrl, zhou2019learn} alleviate this problem by learning through self-exploration. This work describes the implementation of a decentralized motion planning framework using DRL for 4 cooperative robots in 3 different settings.

\section{Problem Formulation}\label{Problem Formulation}

The problem of decentralized motion planning for multi-robot navigation can be formulated as a Partially Observable Markov Decision Process (POMDP). This can be formally described as a 7-tuple $\left \langle S,A,T,R,\Omega,O,\gamma \right \rangle$, where, $S$ is the state space, $A$ is the action space, $T$ is the conditional state transition probability $T\left ( \hat{s}|s,a \right )=P\left ( s_{t+1}=\hat{s} | s_t=s,a_t=a \right )$ that action $a \in A$ in state $s\in S$ at time $t$ will lead to state $\hat{s} \in S$ at time $t+1$, $r=R\left ( s,a \right )$ is the expected immediate reward received after taking action $a$ in current state $s$ and transitioning to new state $\hat{s}$, $\Omega$ is the observation space, $O$ is the conditional observation probability $O\left ( o|\hat{s},a \right )=P\left ( o_{t+1} | s_{t+1}=\hat{s},a_t=a \right )$ of receiving an observation $o \in \Omega$ after transitioning to state $\hat{s}$ due to action $a$ and $\gamma$ is the discount factor satisfying $0 \leq \gamma \leq 1$ which motivates the agent\footnote{Agent is an \textit{intelligent entity} capable of making observations and taking decisions. It can ``learn''.} to either favor immediate expected reward $\left ( \gamma \rightarrow  0 \right )$ or cumulative reward over future horizon $\left ( \gamma \rightarrow  1 \right )$. Thus, the goal of POMDP is to find a policy $\pi$ that makes the agent choose an action $a=\pi\left ( s \right )$ when in state $s$. This reduces the probability distribution $T\left ( \hat{s}|s,a \right )=P\left ( s_{t+1}=\hat{s} | s_t=s,a_t=a \right )$ to $T\left ( \hat{s}|s \right )=P\left ( s_{t+1}=\hat{s} | s_t=s \right )$ (a Markov transition matrix) and the resulting combination behaves like a Markov chain.

\begin{figure}[htbp]
	\centerline{\includegraphics[width=0.5\textwidth]{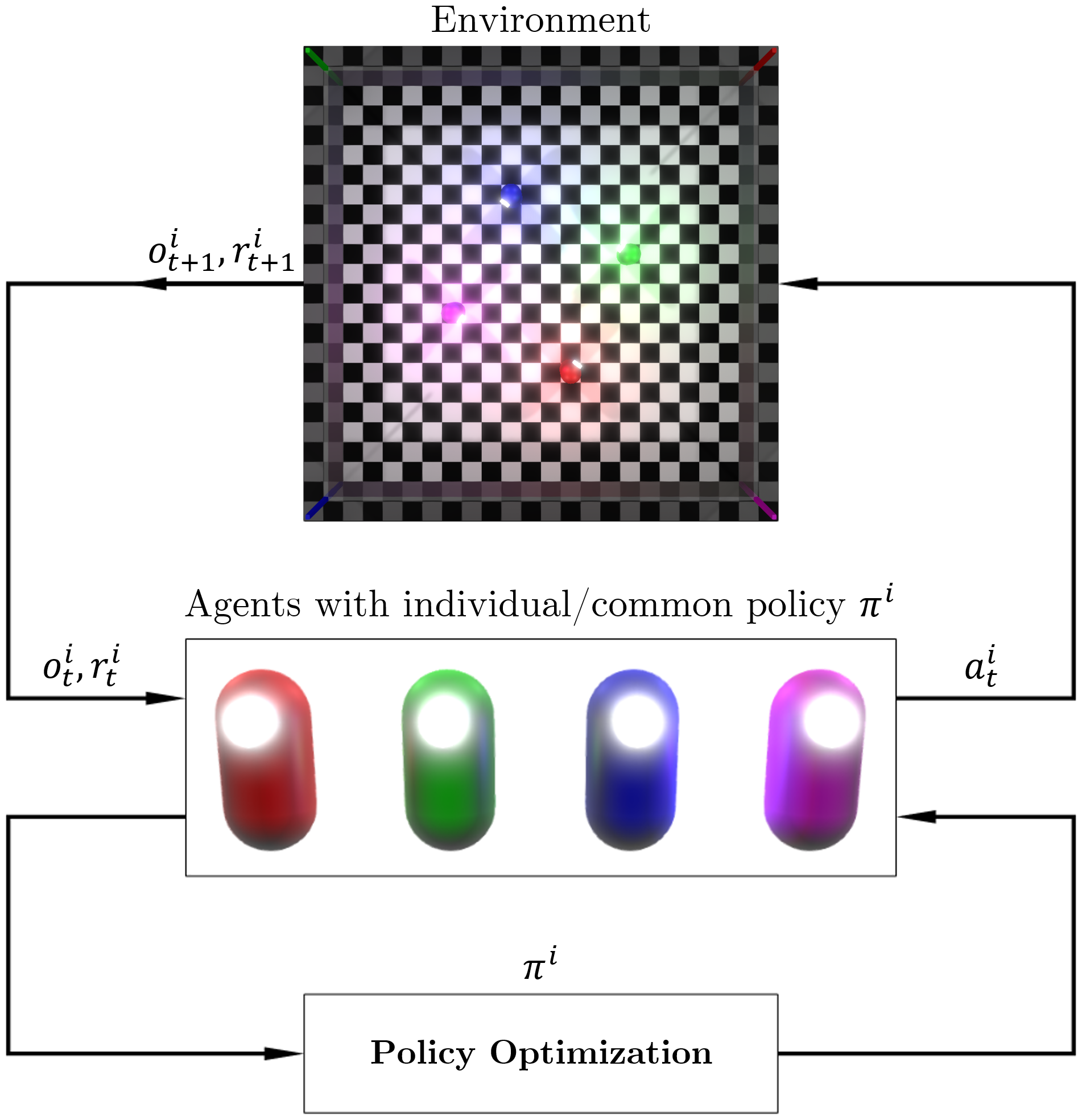}}
	\caption{Learning architecture of a 4-agent system with individual or common policy learning approach. At each timestep $t$, every $i$-th agent gets an observation $o_t^i$ and a reward $r_t^i$ and performs an action $a_t^i$ based on the policy $\pi^i$ learned so far. While this process continues, each policy $\pi^i$ is updated to maximize the expected reward. Each observation-action cycle along with the policy update is referred to as a ``training step''.}
	\label{fig: Learning Architecture}
\end{figure}

\subsection{State Space}\label{State Space}
The state space $S$ is partially observable $s^o \subset S$ and partially hidden $s^h \subset S$. The observable states include robot's position coordinates, $s^o=\left [ p_x, p_z \right ] \in \mathbb{R}^{2}$ while the hidden states include the goal location coordinates, $s^h=\left [ g_x, g_z \right ] \in \mathbb{R}^{2}$. This means that the position coordinates of each agent are observable to its peers while the goal coordinates are hidden (i.e. known to the specific agent itself). Thus, the state vector \eqref{eqn: State Space} of an individual agent can be expressed as,
\begin{align}
s_{t} = \left [ s_{t}^{o}, s_{t}^{h} \right ] \in \mathbb{R}^{4}
\label{eqn: State Space}
\end{align}

\subsection{Observation Space}\label{Observation Space}
This work aims at a physically realizable system. We define the observation space \eqref{eqn: Observation Space} for each of the $N$ agents considering the partially observable state space \eqref{eqn: State Space}. The state observations can be collected using exteroceptive sensing modalities (such as visual or ranging sensors) or through the communication of limited state information between the agents. At each timestep $t$, every $i$-th agent ($0<i<N$) collects a vectorized observation,
\begin{align}
o_{t}^{i} = \left [ p_{t}^{i}, g_{t}^{i}, \tilde{p}_{t}^{i} \right ] \in \mathbb{R}^{4+2(N-1)}
\label{eqn: Observation Space}
\end{align}
where, $p_{t}^{i} = \left [ p_{x}^{i}, p_{z}^{i} \right ]_{t} \in \mathbb{R}^{2}$ is absolute position of $i$-th agent within the environment (X-Z plane\footnote{Unity employs a right-handed coordinate system with Y-axis pointing up.}), $g_{t}^{i} = \left [ g_{x}^{i}-p_{x}^{i}, g_{z}^{i}-p_{z}^{i} \right ]_{t} \in \mathbb{R}^{2}$ is goal location of $i$-th agent relative to itself and $\tilde{p}_{t}^{i} = \left [ \tilde{p}_{x}^{j}-p_{x}^{i}, \tilde{p}_{z}^{j}-p_{z}^{i} \right ]_{t} \in \mathbb{R}^{2(N-1)}$ is position of every other agent $j \in \left [ 0, N-1 \right ]$ relative to the $i$-th agent.

\subsection{Action Space}\label{Action Space}
The actors\footnote{Actor is a \textit{physically present entity} within the environment. It is controlled by an agent. The terms ``actor'' and ``robot'' are interchangeably used in this context.} were generically modeled as non-holonomic differential-drive robots without any specific locomotion mechanism. The continuous space action vector \eqref{eqn: Action Space} for every $i$-th agent comprised of its linear velocity $v$ along local Z-axis and angular velocity $\omega$ about local Y-axis.
\begin{align}
a_{t}^{i} = \left [ v_{z}^{i}, \omega_{y}^{i} \right ]_{t} \in \mathbb{R}^{2}
\label{eqn: Action Space}
\end{align}

The saturation limits of the actuators were also modeled by limiting the control actions $v_{z}^{i} \in \left [ 0, 0.05 \right ]$ and $\omega_{y}^{i} \in \left [ -\pi, \pi \right ]$.

\subsection{Reward Function}\label{Reward Function}
The reward function \eqref{eqn: Reward Function} was crafted such that each agent was awarded $r_{goal}=+20$ upon reaching its goal $g^{i}$ and was penalized by $r_{collision}=-20$ for colliding with its peer agents $\tilde{p}_{t}^{j}; j \in \left [ 0, N-1 \right ]$ of radius $\rho$ or walls $W^{k}; k \in \left [ 0, M \right ]$ of width $w$, after which, a training episode ended.
\begin{align}
r_{t}^{i}  =  
	\begin{cases}
		r_{goal} & \text{if} \quad \left \| g^{i} - p_{t}^{i} \right \| \leqslant 1 \\
		r_{collision} & \text{else if}  \quad \left \| \tilde{p}_{t}^{j} - p_{t}^{i} \right \| \leqslant 2\rho \\
		\; & \text{or} \quad \left \| W^{k} - p_{t}^{i} \right \| \leqslant \rho+\frac{w}{2}
	    \\
		k_{t}*\left \| g^{i} - p_{t}^{i} \right \|^{2} & \text{otherwise}
	\end{cases}
	\label{eqn: Reward Function}
\end{align}

Additionally, the agent was penalized at every timestep with a penalty proportional ($k_{t}=-0.01$) to the square of its distance from the goal. This not only forced the agents to temporally optimize their trajectories, but also accelerated the learning process as the agents were tempted to drive towards their goal in order to minimize this penalty.

\subsection{Optimization Problem}\label{Optimization Problem}
The task of motion planning for $N$ robots can be empirically formulated as an optimization problem aimed at minimizing the expected mean time-to-goal $t_g$ using a policy $\pi$ defined by the parameters $\theta \in \mathbb{R}$:
\begin{align}
\argmin_{\pi_\theta} \quad &\mathbb{E}  \left[\frac{1}{N}\sum_{i=1}^{N} t_g^i | \pi_\theta \right] \label{eqn: Minimization Problem} \\ 
\text{s.t.} \quad & p_t^i = p_{t-1}^i + \Delta t \cdot a_t^i \label{eqn: Kinematic Constraints} \\
\quad & ||p_t^i - \tilde{p}_t^j|| > 2\rho \qquad \forall j \in \left[1, N-1 \right], j \neq i \nonumber \\
\quad & ||p_t^i - W^k|| > \rho + \frac{w}{2} \qquad \forall k \in \left[0, M \right] \label{eqn: Collision Constraint} \\ 
\quad & p_{t_g}^i = g^i \label{eqn: Goal Constraint}
\end{align}
where $a_t \sim \pi_\theta\left ( a_t | o_t \right )$ as stated earlier. The constraints \eqref{eqn: Kinematic Constraints}-\eqref{eqn: Goal Constraint} essentially define a go-to-goal behavior with both static and dynamic collision avoidance -- \eqref{eqn: Kinematic Constraints} is the kinematic constraint governing the state transition $p_{t-1}^i \rightarrow p_t^i$ in a timestep $\Delta t$ due to action $a_t^i$, \eqref{eqn: Collision Constraint} is the collision constraint strictly restricting collision of the ego-robot $i$ of radius $\rho$ with its peers $j$ (also of radius $\rho$) and/or the bounding walls of width $w$, and \eqref{eqn: Goal Constraint} is the goal constraint requiring the $i$-th robot to reach its goal $g^i$ at time instant $t_g$ (i.e. final position is goal location).

In the context of reinforcement learning (Fig. \ref{fig: Learning Architecture}), the constraints \eqref{eqn: Kinematic Constraints}-\eqref{eqn: Goal Constraint} are handled using the reward function defined in \eqref{eqn: Reward Function}, which guides the agents towards optimal behavior. The objective is, therefore,  to learn a policy $\pi_\theta : (s,\tilde{s}^o) \mapsto a$, which maps states of each robot (where $s$ denotes states of the ego-agent, $\tilde{s}^o$ denotes the observable states of the peer agents) to appropriate action $a$, so as to maximize the expected future discounted reward. The time minimization problem defined in \eqref{eqn: Minimization Problem} can now be redefined in terms of a reward maximization problem as follows:
\begin{align}
\argmax_{\pi_\theta \left(s, \, \tilde{s}^{o}\right)} \quad &\mathbb{E}\left [ \sum_{t=0}^{\infty} \gamma^t r_t \right ] \label{eqn: Optimization}
\end{align}

The policy, which maximizes \eqref{eqn: Optimization} is called an \textit{optimal policy} and is denoted as $\pi^*$.

\section{Implementation}\label{Implementation}

In this work, we present the \textit{MARL Simulator}\footnote{\href{https://github.com/Tinker-Twins/MARL-Simulator}{https://github.com/Tinker-Twins/MARL-Simulator}}, a custom simulator developed atop the Unity \cite{unity} game engine and linked with ML-Agents Toolkit \cite{ml-agents}, to train and test Multi-Agent Reinforcement Learning (MARL) behaviors. A sample scenario within the MARL Simulator is shown in Fig. \ref{fig: MARL Simulator}. This simulator is developed with a vision to enable rapid development and testing of DRL-based algorithms aimed at perception, planning and control of multi-robot systems. The fact that this simulation system exploits Unity Editor at its core makes it much more flexible and modular to work with than any other standalone simulators available for such applications. Users can modify existing or develop new environments (with a variety of static and dynamic objects), agents (with customized sensing modalities, learning behaviors and actuation constraints) and features (for logging and analyzing the learned behaviors). Furthermore, virtual prototypes of real robots and environments can be easily modeled in the MARL Simulator and the trained models can be deployed back onto the physical robots for real-time implementation.

\begin{figure}[htbp]
	\centerline{\includegraphics[width=0.5\textwidth]{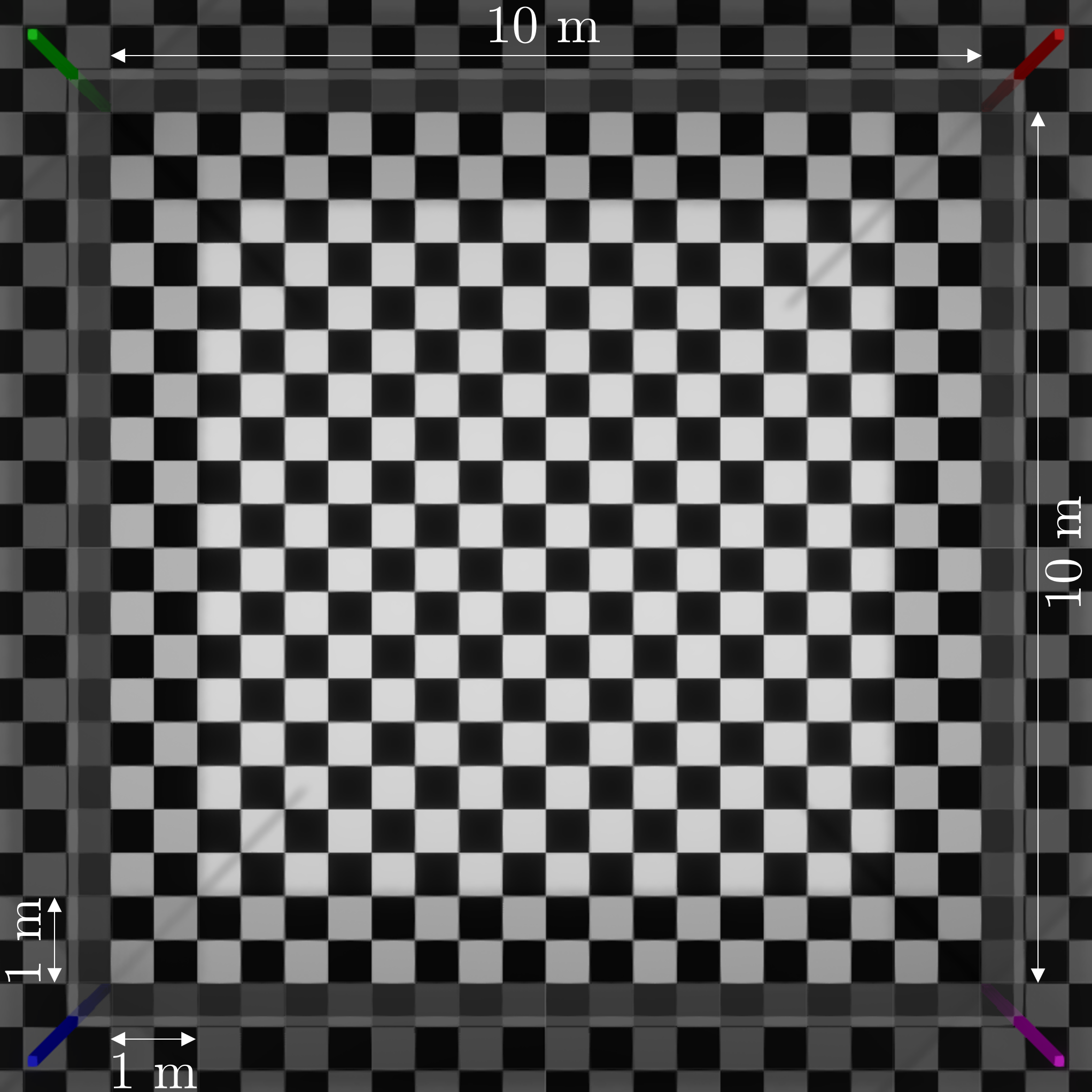}}
	\caption{Square arena within the MARL Simulator.}
	\label{fig: Arena}
\end{figure}

Particularly, this work utilizes the MARL Simulator to train, deploy and analyze multi-robot motion planning and navigation problem in the context of $N = 4$ non-holonomic robots of radius $\rho = 0.25 \; m$ within a $10 \times 10 \; m$ arena defined by walls of width $w = 0.1 \; m$ (Fig. \ref{fig: Arena}). The goal location for each robot was fixed at one of the four corners of the square arena (identified by corner pole of respective color) and it was ensured, at all times (throughout all the experiments), that the robots were spawned only in non-adjacent quadrants w.r.t. their goal locations. We deliberately adopted a sparse observation space while keeping the action space continuous, to challenge the agents by providing them with bare minimal information to accomplish the task of continuous coordinated navigation. Additionally, the decision period was limited to once in 5 steps while allowing the agents to take actions between decisions (like a realistic cyber-physical system). To further add to the complexity of this problem, the agents were left asynchronous and were re-spawned independently after each episode, giving rise to a highly stochastic environment.

\subsection{Experiments}\label{Experiments}
\begin{figure}[htbp]
	\centering
	\subfloat[]{\includegraphics[width=0.16\textwidth]{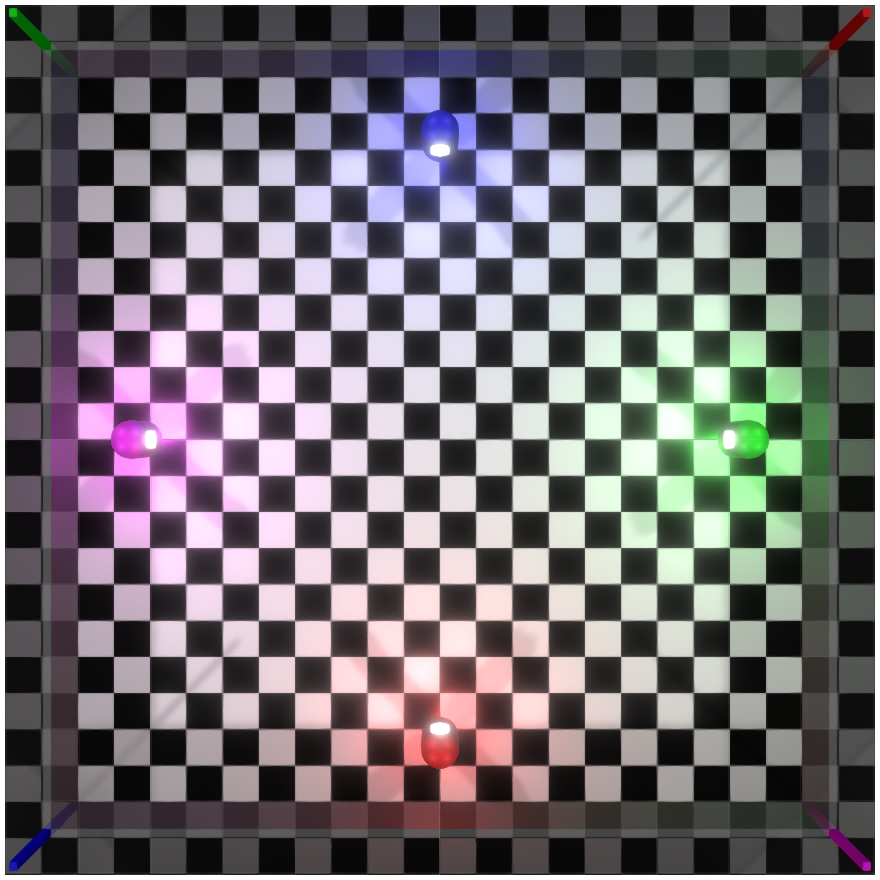}}
	\subfloat[]{\includegraphics[width=0.16\textwidth]{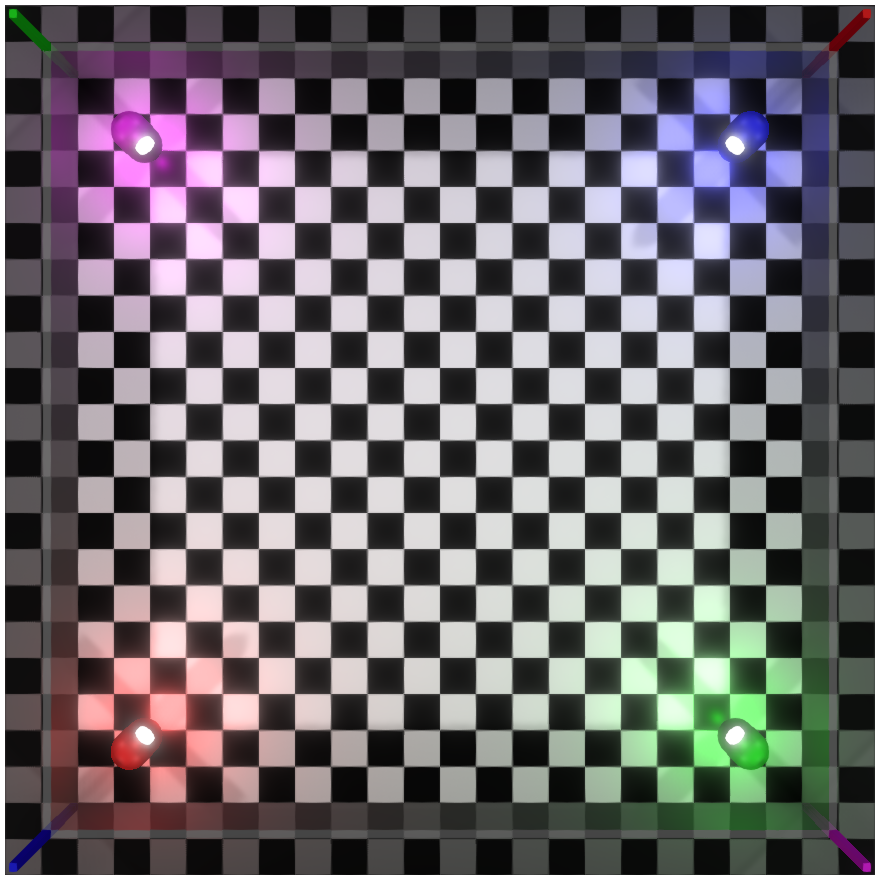}}
	\subfloat[]{\includegraphics[width=0.16\textwidth]{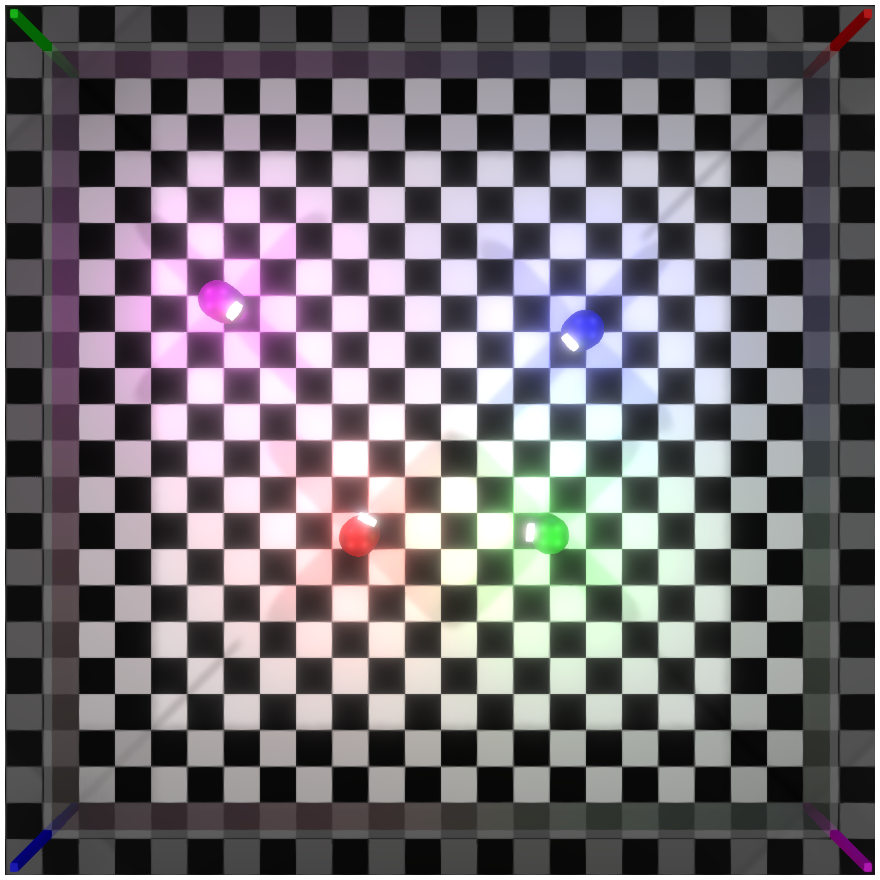}}
	\caption{Episode initialization for a 4-robot system exhibiting (a) G2GCA, (b) APE and (c) G2GCARI\protect\footnotemark behaviors.}
	\label{fig: Experiments}
\end{figure}

A total of 3 experiments were framed, in the order of increasing complexity, to train and deploy the MARL behavior models; the key difference between the three settings being initialization pose of the robots, as illustrated in Fig. \ref{fig: Experiments}. While the first behavior was trained using both individual and common policy approach (to analyze performance in either case), the later were trained using the common policy approach alone, as it was proven to converge faster (w.r.t. time) while yielding similar results (refer to section \ref{Results}).

Following is a summary of the two training approaches:

\begin{itemize}
\item \textbf{Individual Policy (IP):}\\
All agents had the same training configuration; however, each agent was assigned an independent behavior, which forced its policy to be updated solely based on its own experiences.
\item \textbf{Common Policy (CP):}\\
All agents had the same training configuration and a common behavior. A clone of this common policy was attached to each agent, thereby allowing the policy to be updated parallelly based on experiences of all the agents.
\end{itemize}

Following is a summary of the three framed experiments:

\begin{itemize}
\item \textbf{Go-to-Goal with Collision Avoidance (G2GCA):}\\
The robots were spawned at the centers of environment boundaries (i.e. walls), $1 \; m$ distant from the respective wall, facing towards center of the environment. They had to drive towards their goals, which were located at the diagonally right corners w.r.t. each robot, without colliding with each other or the walls.
\item \textbf{Antipodal Exchange (APE):}\\
The robots were spawned at the corners of environment, $(1,1) \; m$ distant from the adjacent walls, facing towards center of the environment. They had to drive towards their goals, which were located at the diagonally opposite corners w.r.t. each robot, without colliding with each other or the walls. Note that spawn location of one robot was the goal location of other (diagonally opposite robot), which made this task challenging.
\item \textbf{G2GCA with Random Initialization (G2GCARI):}\\
The robots were randomly spawned (random position and orientation) within diagonally opposite quadrants w.r.t. their goal locations and had to drive towards their goals without colliding with each other or the walls. This behavior was exceptionally challenging as the robots had to cope up with the extremely stochastic environment.
\end{itemize}

\footnotetext{G2GCARI initializes robots randomly. Fig. \ref{fig: Experiments} shows one such case.}

\subsection{Training Configurations}\label{Training Configurations}
We make use of Proximal Policy Optimization (PPO) \cite{schulman2017proximal} algorithm to train the policies for all the experiments. A Fully Connected Neural Network (FCNN) with input dimensions equal to the observation vector, $\mathbb{R}^{10}$, output dimensions equal to the action vector, $\mathbb{R}^{2}$, and 128 neural units in each of the hidden layers is adopted as a function approximator, $\pi_\theta \left ( a_t | o_t \right )$, which generates an appropriate action $a_t$ given a specific observation $o_t$. Here, the policy parameters $\theta \in \mathbb{R}^d$ are the weights and biases of the neural network. Table \ref{Training Configurations Table} describes the adopted training configurations.

\begin{table}[htbp]
	\caption{Training Configurations for Various Experiments}
	\begin{center}
	\begin{tabular}{|l|c|c|c|}
		\hline
		\multicolumn{1}{|c|}{\multirow{2}{*}{\textbf{Parameter}}} & \multicolumn{3}{c|}{\textbf{Value}}                                         \\ \cline{2-4} 
		\multicolumn{1}{|c|}{}                                    & \textit{\textbf{G2GCA}} & \textit{\textbf{APE}} & \textit{\textbf{G2GCARI}} \\ \hline
		Hidden layers                                             & 2                       & 2                     & 3$^{\mathrm{*}}$          \\ \hline
		Batch size                                                & 1024                    & 1024                  & 1024                      \\ \hline
		Buffer size                                               & 10240                   & 10240                 & 10240                     \\ \hline
		Epochs                                                    & 3                       & 3                     & 3                         \\ \hline
		Learning rate ($\alpha$)                                  & 3e-4                    & 3e-4                  & 3e-4                      \\ \hline
		Learning rate schedule                                    & Linear                  & Linear                & Linear                    \\ \hline
		Entropy regularization strength ($\beta$)                 & 0.05                    & 0.05                  & 0.05                      \\ \hline
		Policy update hyperparameter ($\epsilon$)                 & 0.2                     & 0.2                   & 0.2                       \\ \hline
		Regularization parameter ($\lambda$)                      & 0.97                    & 0.97                  & 0.97                      \\ \hline
		Discount factor ($\gamma$)                                & 0.99                    & 0.99                  & 0.99                      \\ \hline
		Maximum steps                                             & 2.5 M                   & 2.5 M                 & 5.0 M                     \\ \hline
		\multicolumn{4}{l}{$^{\mathrm{*}}$Deeper neural networks induced higher non-linearity to the trajectories.}
	\end{tabular}
	\end{center}
	\label{Training Configurations Table}
\end{table}

\section{Results and Discussion}\label{Results}
\begin{figure*}[htbp]
	\centering{\includegraphics[width=1\textwidth]{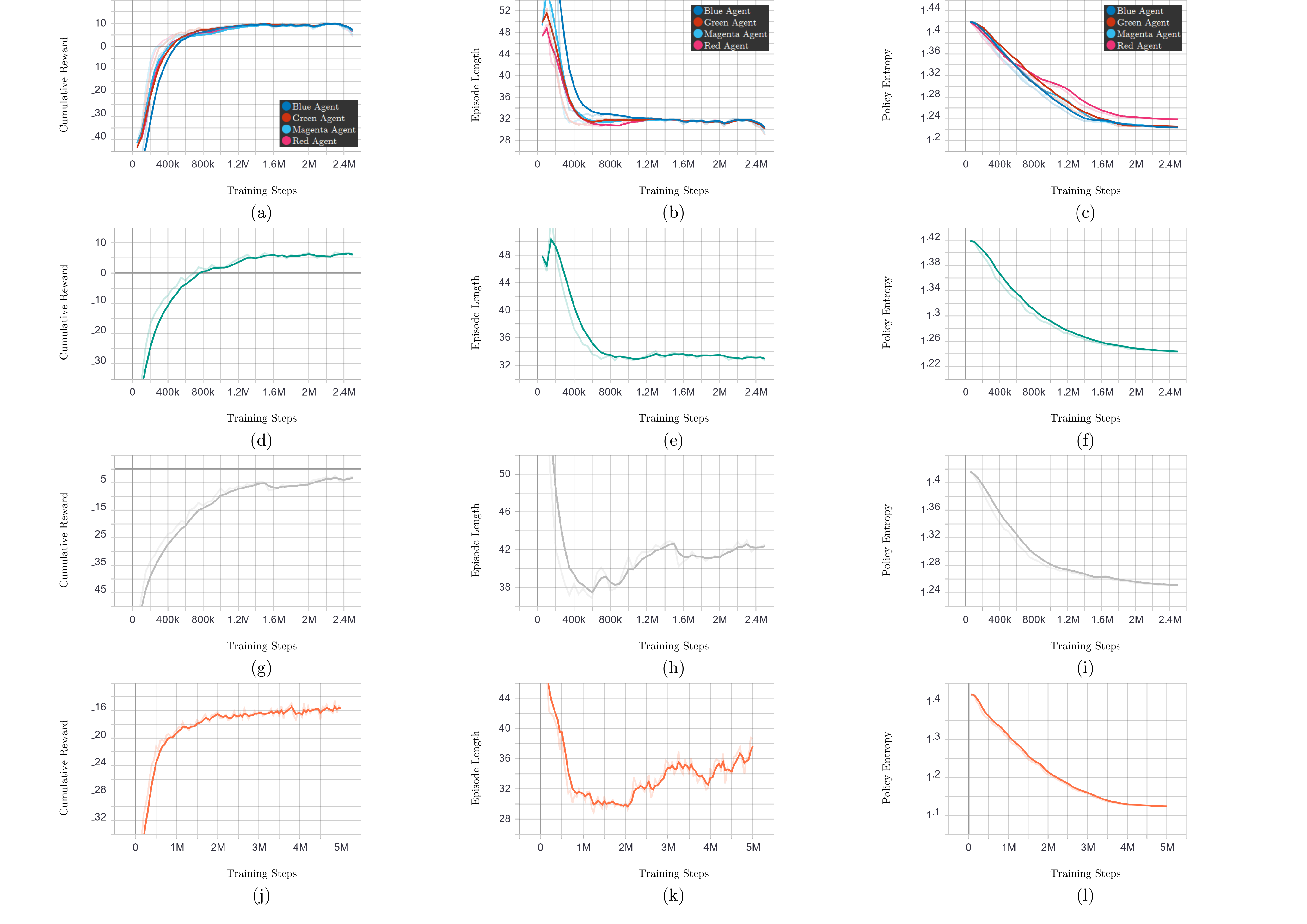}}
	\caption{Training results -- the rows respectively represent G2GCA-IP, G2GCA-CP, APE-CP and G2GCARI-CP experiments and the columns respectively represent cumulative reward, episode length and policy entropy w.r.t. training steps.}
	\label{fig: Training Results}
\end{figure*}

\begin{figure*}[htbp]
	\centering{\includegraphics[width=0.85\textwidth]{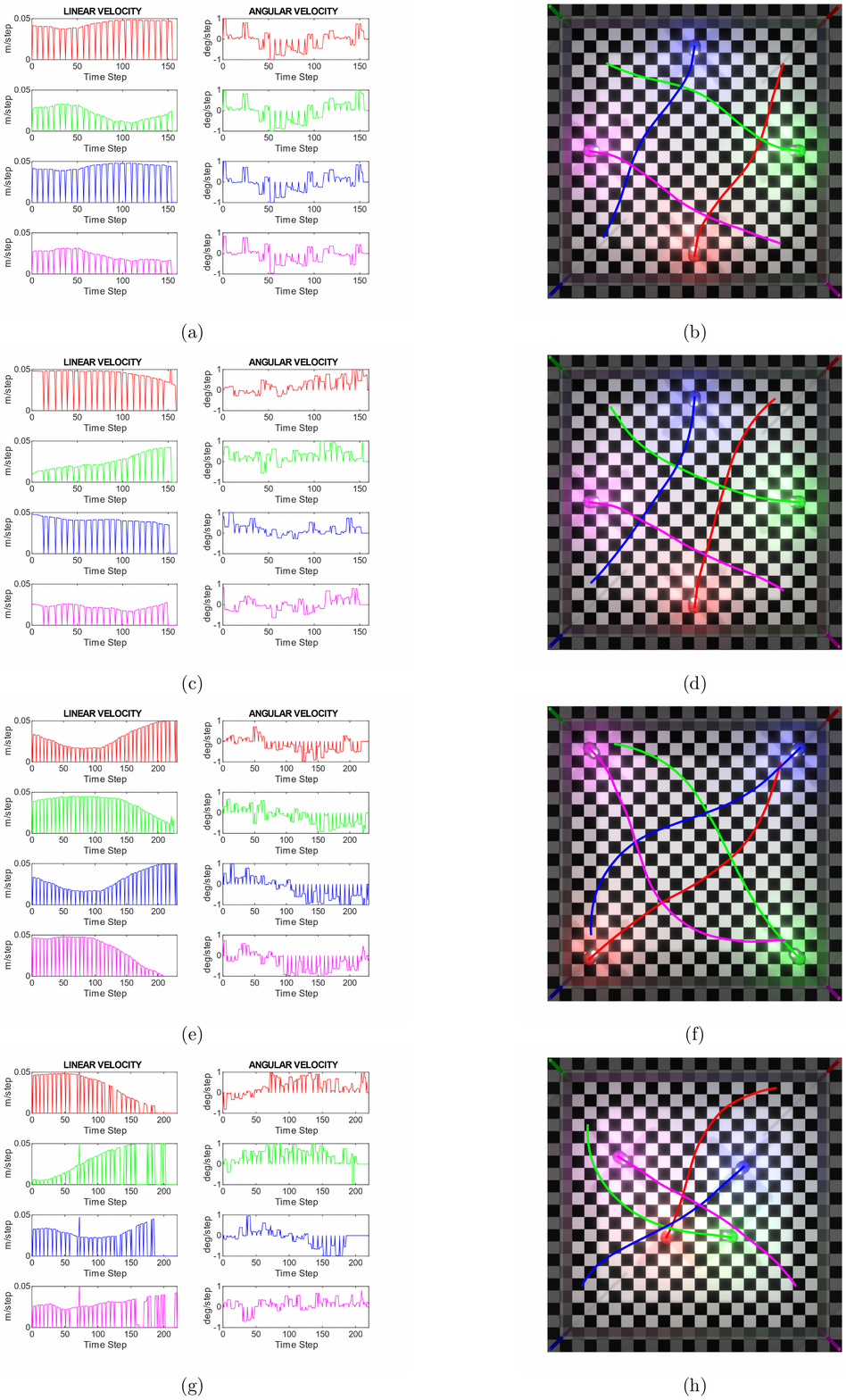}}
	\caption{Deployment results -- the rows respectively represent G2GCA-IP, G2GCA-CP, APE-CP and G2GCARI-CP experiments and the columns respectively represent instantaneous linear and angular velocity plots and trajectory visualizations for each robot.}
	\label{fig: Deployment Results}
\end{figure*}

The training and deployment phases of all the experiments described in section \ref{Experiments} were carried out on a personal computer incorporating Intel i7-8750H CPU, NVIDIA RTX 2070 GPU and 16 GB RAM running Unity 2018.4.24f1 and ML-Agents 0.19.0 with Python 3.7.9 and TensorFlow 2.3.0. While the neural network training and inference phases were both carried out on the CPU, the fact that GPU performed graphical computations of the simulator cannot be disregarded.

\begin{table}[htbp]
	\caption{Training Time for Various Experiments}
	\begin{center}
		\begin{tabular}{|l|l|c|}
			\hline
			\multicolumn{2}{|c|}{\textbf{Experiment}} & \textbf{Training Time (hh:mm:ss)} \\ \hline
			\multirow{4}{*}{G2GCA-IP} & Red Agent     & 12:44:29                          \\ \cline{2-3} 
			& Green Agent   & 12:44:30                          \\ \cline{2-3} 
			& Blue Agent    & 12:44:31                          \\ \cline{2-3} 
			& Magenta Agent & 12:44:32                          \\ \hline
			\multicolumn{2}{|l|}{G2GCA-CP}            & 04:16:21                          \\ \hline
			\multicolumn{2}{|l|}{APE-CP}              & 03:47:17                          \\ \hline
			\multicolumn{2}{|l|}{G2GCARI-CP}          & 08:14:22                          \\ \hline
		\end{tabular}
	\end{center}
	\label{Training Time Table}
\end{table}

Training time for agents in all the experimental settings is summarized in Table \ref{Training Time Table}. It can be observed that G2GCA behavior with individual policy approach took nearly 13 hours to complete 2.5 million training steps as 4 independent policies had to be trained simultaneously, with each agent contributing its experiences to a single policy. The same behavior, when trained using the common policy approach, required only one-third the amount of time since a common policy was being updated based on the experiences of all 4 agents. Since the cumulative reward and success rate was almost similar for the two approaches (refer to section \ref{Training Results} and \ref{Deployment Results}), the common policy approach was followed for the remaining experiments to expedite the training process. The APE behavior took less than 4 hours to complete 2.5 million training steps, since the number of training steps covered per episode was higher owing to the fact that agents had to traverse to diagonally opposite corners, covering more distance. Lastly, the G2GCARI behavior required over 8 hours of training owing to the denser network architecture and twice as many training steps (refer Table \ref{Training Configurations Table}).

The average deployment latency of the proposed approach was 1.22 milliseconds for a single observation-action cycle, while a complete simulation step including other computational components such as rendering, scripts, physics, transformations, illumination, user interface, etc. took about 14 milliseconds. Compared to CO based methods that take somewhere between 10s to 100s of milliseconds \cite{kandhasamy2020scalable} to solve an online optimization problem, our DRL based approach was practically much faster, since the neural networks inherently learned to predict optimal actions given a set of observations in an end-to-end manner without needing to solve highly nonlinear motion models and perform online optimization at each timestep.

\subsection{Training Results}\label{Training Results}
The training results comprise of cumulative reward, episode length and policy entropy plotted against the training steps (Fig. \ref{fig: Training Results}) for all the experiments. A general indication of ``good'' training is that the cumulative reward is maximized and then saturated, the episode length is adequate (longer duration implies agents wandering off in the environment, while very short duration may be an indicative of agents colliding with nearby obstacles) and the policy entropy (i.e. randomness) has decreased steadily as the training progressed.

The agents were able to cash in a maximum reward of about 10 points when trained independently for G2GCA behavior (Fig. \ref{fig: Training Results}a). This was primarily due to the freedom of independent exploration and individual policy training. Towards end of the training, agents could to navigate to their goals within 30 episode steps or 3.20 seconds (Fig. \ref{fig: Training Results}b) as a result of better learning and the policy entropy of three agents converged almost equally to 1.22 by the end of training, indicating that the said agents learned nearly equally; however, red agent exhibited somewhat higher randomness of 1.24 in its action choice (Fig. \ref{fig: Training Results}c) -- this is common in individual policy learning, wherein all policies may not converge at the same time thereby making some agents better than others. Comparatively, the same behavior, when trained using common policy approach, yielded a maximum reward of about 6 points (Fig. \ref{fig: Training Results}d) with a minimum episode length of 33 steps or 3.62 seconds (Fig. \ref{fig: Training Results}e) and final policy entropy of 1.24 (Fig. \ref{fig: Training Results}f); however, the fact that it required substantially less training time (refer Table \ref{Training Time Table}) made common policy approach a better choice.

The agents trained for APE behavior were able to collect a maximum reward of about -4 points (Fig. \ref{fig: Training Results}g), indicating that the time-bound penalty dominated the goal reward (refer section \ref{Reward Function}). The minimum episode length in this case was around 42 steps or 4.83 seconds (Fig. \ref{fig: Training Results}h) owing to the extra distance traversal and the final policy entropy was 1.25 (Fig. \ref{fig: Training Results}i) indicating a well converged policy.

The G2GCARI experiment was highly randomized in terms of agent initialization. The agents could only score a maximum reward of -16 points (Fig. \ref{fig: Training Results}j) owing to the reduced success rate (refer Table \ref{Success Rate Table}). The episode length (Fig. \ref{fig: Training Results}k) was highly fluctuating due to random initialization of the robots near or farther apart from the goal location, but the policy seemed to converge at an entropy value of about 1.125 (Fig. \ref{fig: Training Results}l) due to prolonged training of 5 million steps (refer Table \ref{Training Configurations Table}).

\subsection{Deployment Results}\label{Deployment Results}
This work was particularly focused on testing the robustness of the trained models towards the stochastic nature of the environment. The trained agents were therefore tested for their performance in highly stochastic scenarios. Table \ref{Success Rate Table} holds the success rates for all the experiments. These values are based on 500 random episodes, wherein the agents were initially spawned collectively, but the re-spawning after each episode was left independent to each agent so as to create a highly stochastic setting (especially after $\sim$100 episodes).

The variation in success rates (78.35\% versus 88.9\%) for G2GCA behavior trained using individual and common policy approaches, respectively, is merely due to the difference in the relative initialization of the robots during various training and/or testing episodes. The success rate of APE behavior (89.7\%) is nearly equal to that of G2GCA behavior (88.9\%), indicating almost equal mastery in both the cases, even though the APE behavior was anticipated to be tougher than G2GCA. Finally, the success rate of G2GCARI behavior is lower than others (36.5\%) due to two possible reasons: (i) insufficient generalization of the neural network policy to the highly randomized setting, and/or (ii) unjust inter-agent collision resulting from agents spawning too close to one another.
\begin{table}[htbp]
	\caption{Success Rates for Various Experiments\protect\footnotemark}
	\begin{center}
		\begin{tabular}{|l|c|c|c|c|}
			\hline
			\multicolumn{1}{|c|}{\textbf{Agent}} & \textbf{G2GCA-IP}         & \textbf{G2GCA-CP}        & \textbf{APE-CP}           & \textbf{G2GCARI-CP}      \\ \hline
			Red                             & 77.4\%                    & 84.8\%                   & 89.6\%                    & 38.4\%                   \\ \hline
			Green                           & 78.6\%                    & 88.6\%                   & 89.6\%                    & 37.0\%                   \\ \hline
			Blue                            & 79.4\%                    & 92.2\%                   & 90.8\%                    & 34.8\%                   \\ \hline
			Magenta                         & 78.0\%                    & 90.0\%                   & 88.8\%                    & 35.8\%                   \\ \hline
			\textit{\textbf{Average}}             & \textit{\textbf{78.35\%}} & \textit{\textbf{88.9\%}} & \textit{\textbf{89.7\%}} & \textit{\textbf{36.5\%}} \\ \hline
		\end{tabular}
	\end{center}
	\label{Success Rate Table}
\end{table}
\footnotetext{Video available at \href{https://youtu.be/BIb9FIvENaU}{https://youtu.be/BIb9FIvENaU}}

In addition to reporting success rates as a metric of meeting the safe navigation objective, the trajectories of all agents along with their motion control commands (Fig. \ref{fig: Deployment Results}) were also analyzed in order to infer the efficiency of motion planning. It is evident that the agents planned their paths quite optimally not only to avoid colliding with each other or the walls, but also to minimize the time-to-goal. It is also evident that each agent planned its motion independently, i.e. in a completely decentralized manner.

Fig. \ref{fig: Deployment Results}a and \ref{fig: Deployment Results}b illustrate the linear and angular velocities of the robots along with their trajectory plots in the G2GCA-IP setting. It can be observed that all the agents followed similar trajectory profiles, however, their strategies for trajectory optimization were not same. While red and blue agents chose to maximize their linear velocities for achieving temporal optimization, green and magenta agents smartly traveled less distance at moderately high speeds to achieve similar time-to-goal. Furthermore, the variations in velocity profiles also indicate the cautious motion planning in closer proximity of peer agents.

In case of G2GCA-CP experiment (Fig. \ref{fig: Deployment Results}c and \ref{fig: Deployment Results}d), the agents were observed to take quite different trajectory profiles as compared to G2GCA-IP setting. Results indicate that the agents preferred to follow concave trajectories as opposed to convex ones, with higher smoothness and lower non-linearity (as can be observed from the trajectory visualization and inferred from the angular velocity plot). A common observation in case of G2GCA-CP and G2GCA-IP experiments is the strategy used by the agents for optimizing time-to-goal (as described earlier). In both the settings, agents took almost 150 time steps to successfully navigate to their respective goal locations whilst following all motion constraints \eqref{eqn: Kinematic Constraints}-\eqref{eqn: Goal Constraint}.

The motion analysis results for APE test scenario (Fig. \ref{fig: Deployment Results}e and \ref{fig: Deployment Results}f) primarily revealed that diagonally opposite agents had learned to optimize their trajectories in a similar fashion. While red and blue agents started with a moderate velocity, slowed down a bit and then accelerated to maximum velocity, green and magenta agents did the exact opposite by starting fast and then slowing down when proximal to the goal location. This is quite an intelligent strategy to reduce the probability of collision with peer agents, yet optimizing time-to-goal. It is to be noted that in certain situations, the agents preferred taking a slightly longer route to avoid collision with peers, but covered up the increased distance with a higher velocity so as to avoid increasing time-to-goal (e.g. second half of the trajectory followed by the blue agent). It is also worth noting that the agents took about 200 time steps to reach goal destinations as opposed to 150 steps in case of G2GCA-CP and G2GCA-IP experiments since antipodal exchange requires the agents to cover a longer distance from spawning location to goal.

Finally, a sample episode from the G2GCARI setting (Fig. \ref{fig: Deployment Results}g and \ref{fig: Deployment Results}h), depicts each agent having a different distance-to-goal owing to the fact that all agents were spawned at random locations with random orientations. As a result, the motion planned by each agent was uniquely subject to its own state, observable states of its peer agents and its relative goal location. A particularly impressive observation here is that since the red and green agents were spawned quite close, the red agent proceeded quickly with high velocity while the green agent accelerated gradually (cautiously with lesser velocity) to avoid colliding with the red agent, and later increased its velocity to reach the goal location quickly. This episode also attests the spatial trajectory optimization claim, as is evident from the trajectory visualization -- the robots follow tight curves around collision-prone zones without deviating much from the straight-line path (especially for the blue and magenta agents).

\section{Conclusion}\label{Conclusion}
This work was aimed at training a group of non-holonomic robots to navigate to their goals while optimizing their trajectories both spatially and temporally. While spatial optimization was concerned with collision avoidance, temporal optimization was concerned with minimizing the time-to-goal. A total of four experiments were carried out to investigate the task of coordinated motion planning for 4 cooperative agents. Initially, the common and individual policy learning approaches were compared in the G2GCA experiment, and the faster-converging common policy approach was followed for the later experiments, namely APE and G2GCARI. The agents were successfully able to learn the complex task of decentralized motion planning in an end-to-end manner by making use of limited observations, even in a randomized setting (G2GCARI). This work shall be pursued further to investigate the navigation problem of multiple non-cooperative agents.

\bibliographystyle{IEEEtran}
\bibliography{Bibliography}

\end{document}